\pdfoutput=1

\documentclass[twocolumn,fleqn,10pt]{wlscirep}
\usepackage{enumitem}
\usepackage[utf8]{inputenc}
\usepackage[T1]{fontenc}
\usepackage{color}
\usepackage{flushend}
\usepackage{multirow}
\usepackage{pifont}
\usepackage[ruled,vlined]{algorithm2e}
\usepackage{pdfpages}
\SetKwInput{KwInput}{Input}
\SetKwInput{KwOutput}{Output}
\usepackage{amsthm}
\usepackage{framed}
\usepackage{mathrsfs}

\newtheoremstyle{ncomm_theorem}
{}
{}
{}
{}
{}
{}
{}
{\thmname{\textit{#1}}\thmnumber{ \textit{#2}: }\thmnote{#3}}

\theoremstyle{ncomm_theorem}

\usepackage{bm}
\usepackage{xspace}

\unboldmath

\title{\vspace{25pt}Automated Architecture Search for Brain-inspired Hyperdimensional Computing}

\author[1]{Junhuan Yang}
\author[2]{Yi Sheng}
\author[3]{Sizhe Zhang}
\author[3]{Ruixuan Wang}
\author[4]{Kenneth Foreman}
\author[4]{Mikell Paige}
\author[3]{Xun Jiao}
\author[2]{Weiwen Jiang}
\author[1]{Lei Yang}
\affil[*]{\textbf{(Preprint Version)}}
\affil[1]{Department of Electrical and Computer Engineering, University of New Mexico, Albuquerque, NM 87131, USA}
\affil[2]{Department of Electrical and Computer Engineering, George Mason University, Fairfax, VA 22030, USA}
\affil[3]{Department of Electrical and Computer Engineering, Villanova University, Villanova, PA 19085, USA}
\affil[4]{Department of Chemistry and Biochemistry, George Mason University, Fairfax, VA 22030, USA}

\begin{abstract}
This paper represents the first effort to explore an automated architecture search for hyperdimensional computing (HDC), a type of brain-inspired neural network. Currently, HDC design is largely carried out in an application-specific ad-hoc manner, which significantly limits its application. Furthermore, the approach leads to inferior accuracy and efficiency, which suggests that HDC cannot perform competitively against deep neural networks. Herein, we present a thorough study to formulate an HDC architecture search space. On top of the search space, we apply reinforcement-learning to automatically explore the HDC architectures. The searched HDC architectures show competitive performance on case studies involving a drug discovery dataset and a language recognition task. 
On the Clintox dataset, which tries to learn features from developed drugs that passed/failed clinical trials for toxicity reasons, the searched HDC architecture obtains the state-of-the-art ROC-AUC scores, which are 0.80\% higher than the manually designed HDC and 9.75\% higher than conventional neural networks.
Similar results are achieved on the language recognition task, with 1.27\% higher performance than conventional methods.

\end{abstract}
\begin{document}

\setlength{\textfloatsep}{3pt}
\setlength{\floatsep}{3pt}
\setlength{\dbltextfloatsep}{3pt}
\setlength{\abovecaptionskip}{3pt}

\flushbottom
\maketitle

\section{Introduction} \label{sec:Intro}

Machine learning (ML) has been widely applied in different domains, such as computer vision \cite{xu2021computer}, language process \cite{cai2021natural}, medical imaging \cite{barragan2021artificial}, and drug discovery \cite{elbadawi2021advanced}. 
Recently, brain-inspired hyperdimensional computing (a.k.a., HDC) has demonstrated its superiority in ML tasks.
The fundamental of HDC is motivated by the observation that the key aspects of human memory, perception, and cognition can be explained by the mathematical properties of high-dimensional spaces \cite{imani2020machine}.
Specifically, HDC utilizes a high-dimensional space, called hypervectors (HVs), to perform calculations.
Different features in the data will be represented as orthogonal vectors, and operations will be applied for specific tasks (e.g., generate a set of vectors to represent classes in classification tasks).
In order to create the orthogonal vectors, a typical way is to randomly generate HVs with high dimensionality, such as 10,000 \cite{ge2020classification}. 
Benefiting from the high dimensionality of fundamental computing vectors, HDC has the advantage of being robust.
In comparison to conventional neural networks, the regular operations in HDC can enable high parallelism and reduced storage requirement.
As such, HDC is both scalable and energy-efficient with low-latency in both training and inference \cite{ge2020classification}.

The current HDC architecture, however, is mainly designed in an application-specific ad-hoc manner, which greatly limits its generality \cite{khan2021brain,duan2021hdcog}. Furthermore, the fundamental question of what is the best HDC architecture has not yet been answered.
Targeting different domain applications, ML algorithms need to be customized to maximize performance and efficiency. In conventional neural network based ML, neural architecture search (NAS) is an automatic technique to design neural networks, which has achieved great success in designing effective neural architectures \cite{elsken2019neural}. 
When it comes to HDC, applying the NAS technique to design HDC architecture is apparently straightforward; however, the newness of the computing paradigm of HDC brings challenges in applying NAS for HDC.
More specifically, it is even unclear what search space of the HDC architecture NAS can explore.

To the best of our knowledge, this is the first work to conduct an automated architecture search for HDC.
We proposed a holistic framework, namely \textbf{AutoHDC}, to carry out the exploration.
By a throughout analysis of all operations in HDC, AutoHDC first constructs the fundamental search space in the design of the HDC architecture. 
On top of the search space, a reinforcement learning based search loop is devised to automatically search for the best HDC architecture for a given application.

We have out a case study on drug discovery tasks and language processing tasks to evaluate the effectiveness of the proposed AutoHDC framework.
For the drug discovery dataset, we utilized the Simplified Molecular-Input Line-Entry System (SMILES) to encode the molecule as a sequence of data; while the language data remains in sequential format.
AutoHDC generates the tailored HDC architecture for different tasks.
Results show that AutoHDC can achieve state-of-the-art performance on both datasets.
More specifically, on the Clintox dataset, HDC tries to learn features from developed drugs that passed/failed clinical trials for toxicity reasons. The searched HDC architecture obtained the state-of-the-art ROC-AUC scores, which are 0.80\% higher than the manually designed HDC and 9.75\% higher than the conventional neural networks.
On the language recognition task, AutoHDC shows a 1.27\% accuracy improvement compared with the conventional methods.

The main contributions of this paper are as follows.
\begin{itemize}[noitemsep,topsep=0pt,parsep=0pt,partopsep=0pt]
    \item To the best of our knowledge, this is the first work to explore the architecture of hyperdimensional computing (HDC).
    A holistic framework, namely AutoHDC, is proposed to generate the optimal HDC architecture for a given dataset.
    \item AutoHDC formulates the search space of HDC architecture, which includes the variable sparsity (percentage of elements) of hypervectors, rotation (permutations) of different digits, and performing different element-wise operations during the training process.
    \item We have conducted a case study on a drug discovery dataset and a language recognition task to evaluate AutoHDC. Experimental results show that AutoHDC can identify models for different datasets with state-of-the-art performance.
\end{itemize}




For the remainder of the paper: Section 2 reviews related background and provides our motivation. Section 3 presents our AutoHDC  framework. Section 4 shows the experimental results and Section 5 gives our concluding remarks.

\section{Related work and Motivation}\label{sec:pre}

Data representation, transformation, and retrieval play an important role in computing systems. \cite{ge2020classification}
Hypervectors, the foundation of HDC, are high-dimensional (usually in the thousands) and generated with (pseudo-)random and i.i.d. elements \cite{kanerva2009hyperdimensional}. For an HV with $d$ dimension, we can denote it as equation \ref{equ:hv}, where $\overrightarrow{H_i}$ represents the $i-th$ elements in HV $\overrightarrow{H}$. It is used to represent data, which is different from dealing with bits in
classical computing. But what is the best dimension for HVs, and can the dimensions of HVs be smaller than 10,000? HVs with a large dimensions can bring robustness, along with more computation. In contrast, HVs with a smaller dimensions may lose robustness but bring speed. Basaklar and co-workers \cite{basaklar2021hypervector} proposed to reduce dimensionality greatly from 8192 to 64 with little loss in accuracy. However, manual design of the HVs was required, which is obviously unsuitable for most alternate scenarios. Additionally, the manual design of high-dimensional HVs can be very challenging or even impossible to achieve. Therefore, an automated design of a suitable dimension for different applications is attractive.

\begin{equation}
    \centering
    \overrightarrow{H} \  = \ (\overrightarrow{H_1}, \ \overrightarrow{H_2}, \ ..., \ \overrightarrow{H_i}, \ ..., \ \overrightarrow{H_d}) \label{equ:hv}
\end{equation}

\begin{figure}[t]
\begin{center}
\centerline{\includegraphics[width=\columnwidth]{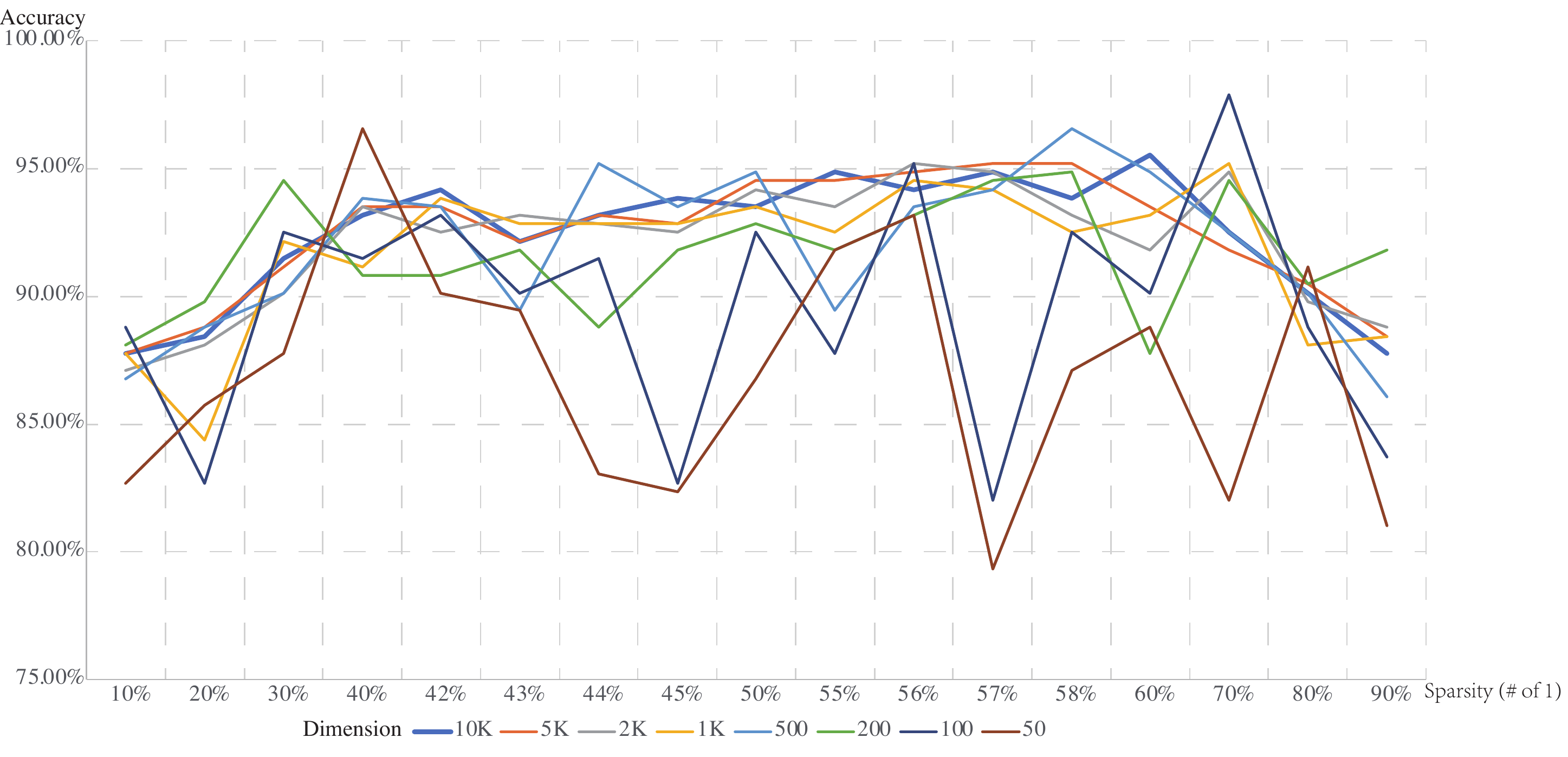}}
\vskip 0.1in
\caption{HDC performance on Clintox with different dimensions and sparsity HVs}
\label{fig:motivation}
\end{center}
\vskip -0.2in
\end{figure}

Another issue of (pseudo-)randomly generated HVs is the sparsity of HVs' elements. For example, the number of \textit{1} and \textit{0} in an HV with dimension 10,000 should be nearly 5,000 and 5,000 respectively if the HV's datatype is binary. But why should the ratio of \textit{1} and \textit{0} both be 50\%? The HVs are employed to represent the data. But the static sparsity of elements may not be perfect for all applications or data. To verify this, we did a set of experiments to explore the impact on the performance of HDC from various dimensions and sparsity. Figure \ref{fig:motivation} shows the accuracy of HVs with different dimensions and sparsity on the dataset Clintox. The best accuracy comes from a bipolar HV with intermediate dimensionality, which is greater than 50\% sparsity. Results indicate that finding a suitable dimensionality and sparsity of HVs for an HDC must be considered.

Datatype is another property to consider. During the whole process, 3 kinds of HVs will be produced. The first one is (pseudo-)randomly generated HVs, and its datatype can be binary, integer, real, or complex \cite{hersche2018exploring}. The other two kinds of HVs will be calculated through the process. Current HDC may not specify the datatype of these two kinds of HVs. However, the datatype may also affect the training and inference result. This is a critical factor for HDC model design.

Inspired by the dimension, sparsity, and datatype of HVs, other properties and operations during the HDC process can be varied and they may also improve HDC performance. To transform data, the current HDC performs three operations: multiplication, addition, and permutation \cite{ge2020classification}. In current implementations of HDC, some researchers also use $XOR$ to replace multiplication. Both of them are 
used to associate two HVs. Besides multiplication and $XOR$, we can also use other operations to associate two HVs. 
The permutation is also a choice in HDC. In some applications, the authors choose to do permutation\cite{karunaratne2020memory}, but there is another choice to compute without permutation. The permutation is used to represent the sequence of data appearance in HDC. Since the HVs are generated randomly, the shift of permutation can bring different results.

Based on these observations, we are motivated to design suitable HDC models, in order to find the suitable HVs and relative encoding processes and computations for different datasets or applications. For current deep neural networks, NAS is used for designing neural networks automatically. Thus, we are inspired to design a framework to search dedicated HDC models automatically for various applications and datasets.

\section{Framework of Automated Architecture Search for Hyperdimensional Computing }\label{sec:frame}
In this section, we will present the proposed framework together with a detailed  exploration of a suitable hyperdimensional computing architecture for different datasets.

\begin{figure}[t]
\begin{center}
\centerline{\includegraphics[width=\columnwidth]{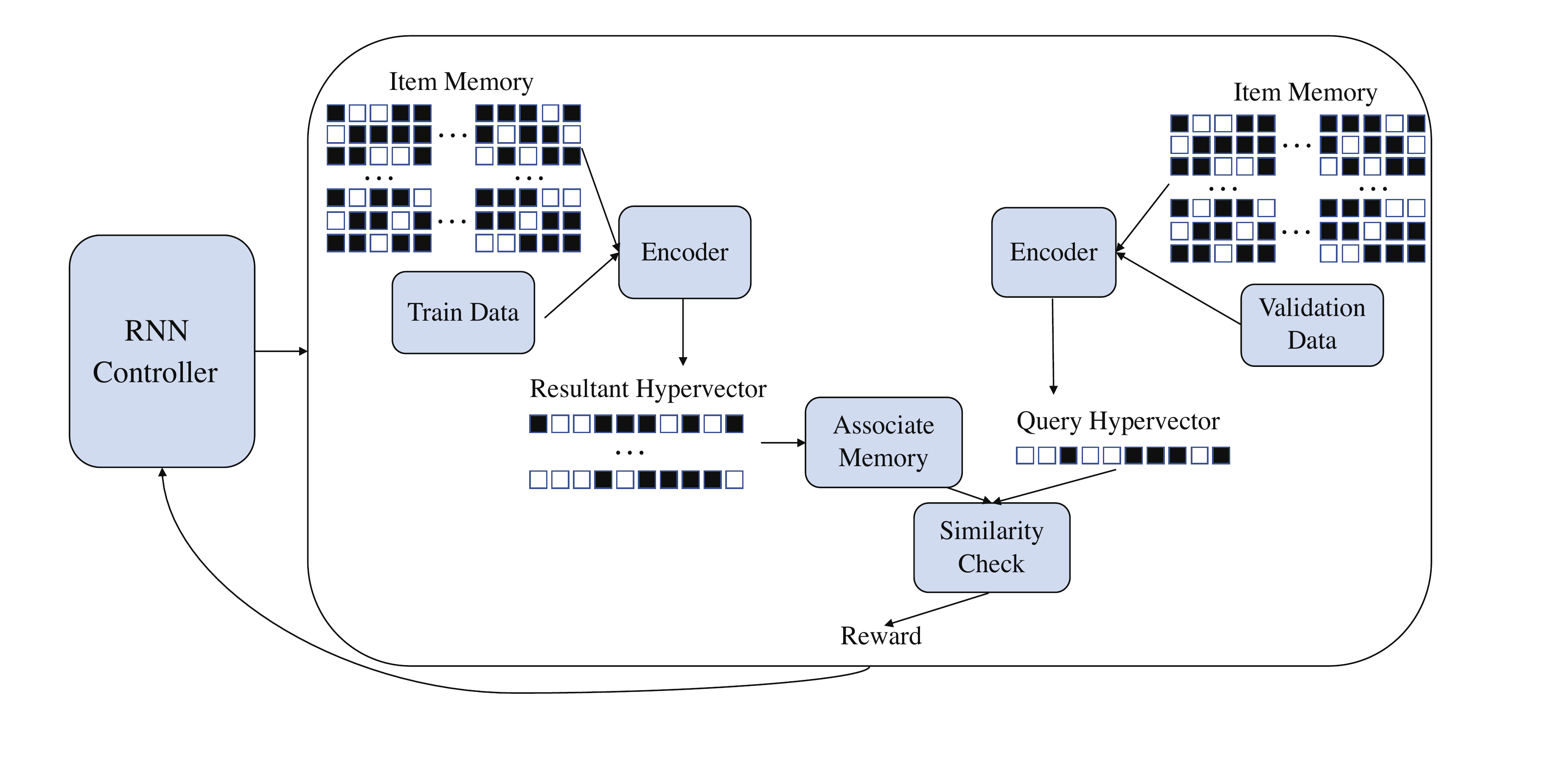}}
\vskip 0.1in
\caption{An overview of hyperdimensional computing architecture search framework.}
\label{farme_overview}
\end{center}
\end{figure}

\subsection{Framework Overview}
Figure \ref{farme_overview} shows an overview of the hyperdimensional computing architecture search framework.
The framework contains an RNN based controller and an HDC executor to produce a reward for a searched architecture of a specific dataset.

Tokenization is the first step, which converts text into corresponding sets of numerical tokens. It consists of 3 procedures: converting the strings (e.g., SMILES string) into tokens and then assigning numbers for the tokens according to their appearance times. Finally, the text is transferred to corresponding numerical tokens. All the texts in the dataset are treated as textual strings and every unique character in the string will be assigned a unique number to form the numerical tokens.

The randomly generated base hypervectors which use a seed are stored in the item memory. The item memory contains \textit{base hypervectors} ($\overrightarrow{B}$) with the same number of keys in the token dictionary. As the tokenization carried out, every numerical token has a related base hypervector. We denote the item memory as \textbf{B} = \{$\overrightarrow{B_1}$, $\overrightarrow{B_2}$, ..., $\overrightarrow{B_i}$, ..., $\overrightarrow{B_m}$\} where the $\overrightarrow{B_i}$ represents the base hypervector with index $i$ and $m$ represents the number of  keys in the token dictionary. Thus, in the encoder, every string in the training data and validation data can be encoded into a set of hypervectors.

For the training and validation data, after encoding, we refer to these hypervectors for unit strings (e.g., SMILES) as \textit{encoded hypervectors}, denoted as $\overrightarrow{ET}$ and $\overrightarrow{EV}$.  $\overrightarrow{ET}$ will be used in the training and retraining processes, while $\overrightarrow{EV}$ will be used to do inference to validate the performance of the model. Training is the process of establishing the original associative memory which includes specific hypervectors for each class relative to the training data. We call the class hypervectors in associative memory \textit{resultant hypervectors}, denoted as \textbf{R} = \{$\overrightarrow{R_1}$, $\overrightarrow{R_2}$, ..., $\overrightarrow{R_j}$, ..., $\overrightarrow{R_c}$\}, where $j$ and $c$ represent the class label. Each training encoded hypervectors $\overrightarrow{ET_k}$ are added to the corresponding resultant hypervectors $\overrightarrow{R_j}$ according to its label.

Note that, the original associative memory after training is sufficient for some scenarios, however, retraining can help improve the HDC performance. Retraining is a process to fine-tune the resultant hypervectors using the encoded hypervectors $\overrightarrow{ET}$. The original associative memory is used to predict the corresponding label of $\overrightarrow{ET_k}$. If the prediction is not correct, the relative resultant hypervector does not show or does not contain the correct information. Thus, the resultant hypervector with the prediction label should subtract the $\overrightarrow{ET_k}$, and the resultant hypervector with the correct label should be added with $\overrightarrow{ET_k}$.

The validation data is used to evaluate the performance of the model after training and retraining. The process of encoding validation data to encoded hypervectors $\overrightarrow{EV}$ is the same as the process of encoding training data. The encoded hypervector is also called a query hypervector ($\overrightarrow{Q}$). The hamming similarity or cosine similarity is then calculated between a  query hypervector $\overrightarrow{Q_l}$ and each resultant hypervector $\overrightarrow{R_j}$ in associative memory.  The result of the largest similarity, such as from $\overrightarrow{Q_l}$ and $\overrightarrow{R_j}$, shows that $\overrightarrow{R_j}$ is the most similar resultant hypervector of the query hypervector $\overrightarrow{Q_l}$, and that the model which predicts the query hypervector $\overrightarrow{Q_l}$ is most likely to have the same label as that of the resultant hypervector $\overrightarrow{R_j}$. 

According to the demands, we can design specific reward calculations to give back to the RNN controller. For example, we may use  accuracy as a reward if we want to search for a model with high accuracy.

\subsection{Executor process}

\begin{figure}[t]
\begin{center}
\centerline{\includegraphics[width=\columnwidth]{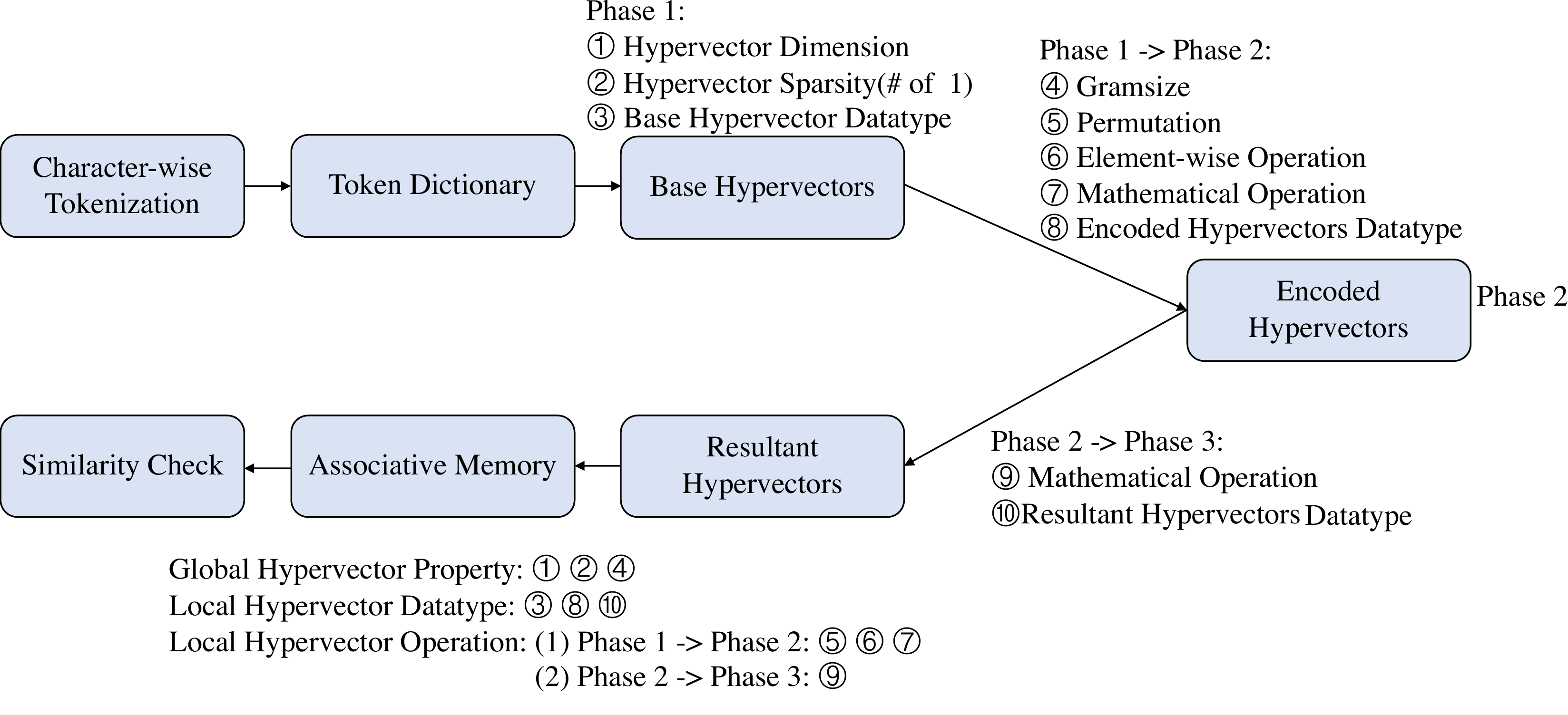}}
\vskip 0.1in
\caption{HDC Executor process}
\label{search_overview}
\end{center}
\vskip -0.2in
\end{figure}

The subsection above has demonstrated the preliminaries on HDC and an overview of our hyperdimensional computing architecture search framework. In this subsection, we will illustrate the details of our hyperdimensional computing architecture search.

\begin{figure*}[!t]
\normalsize
\begin{equation}
\overrightarrow{H_P} \ \  * \ \  \overrightarrow{H_Q} \ \ = \ \ 
\{\overrightarrow{H_p{}_({}_1{}_)}\ \  * \ \  \overrightarrow{H_q{}_({}_1{}_)},\ \  \overrightarrow{H_p{}_({}_2{}_)} \ \  * \ \ \overrightarrow{H_q{}_({}_2{}_)}, \ \  ...,\ \  \overrightarrow{H_p{}_({}_i{}_)} \ \  * \ \  \overrightarrow{H_q{}_({}_i{}_)}, \ \  ...,\ \  \overrightarrow{H_p{}_({}_d{}_)} \ \  * \ \ \overrightarrow{H_q{}_({}_d{}_)}\} \label{equ:mul}
\end{equation}
\begin{equation}
\overrightarrow{H_P} \ \  XOR \ \  \overrightarrow{H_Q} \ \ = \ \ 
\{\overrightarrow{H_p{}_({}_1{}_)} \ \  XOR \ \  \overrightarrow{H_q{}_({}_1{}_)}, \ \  \overrightarrow{H_p{}_({}_2{}_)} \ \  XOR \ \ \overrightarrow{H_q{}_({}_2{}_)},\ \ ..., \ \  \overrightarrow{H_p{}_({}_i{}_)} \ \  XOR \ \  \overrightarrow{H_q{}_({}_i{}_)},\ \   ...,\ \  \overrightarrow{H_p{}_({}_d{}_)} \ \  XOR \ \  \overrightarrow{H_q{}_({}_d{}_)}\} \label{equ:xor}
\end{equation}
\begin{equation}
\overrightarrow{H_P} \ \  AND \ \  \overrightarrow{H_Q} \ \ = \ \ 
\{\overrightarrow{H_p{}_({}_1{}_)} \ \  AND \ \  \overrightarrow{H_q{}_({}_1{}_)},\ \  \overrightarrow{H_p{}_({}_2{}_)} \ \  AND \ \ \overrightarrow{H_q{}_({}_2{}_)},  \ \ ...,\ \  \overrightarrow{H_p{}_({}_i{}_)} \ \  AND \ \  \overrightarrow{H_q{}_({}_i{}_)}, \ \ ...,\ \  \overrightarrow{H_p{}_({}_d{}_)} \ \  AND \ \  \overrightarrow{H_q{}_({}_d{}_)}\} \label{equ:and}
\end{equation}
\begin{equation}
\overrightarrow{H_P} \ \  OR \ \  \overrightarrow{H_Q} \ \ = \ \ 
\{\overrightarrow{H_p{}_({}_1{}_)} \ \  OR \ \  \overrightarrow{H_q{}_({}_1{}_)},\ \  \overrightarrow{H_p{}_({}_2{}_)} \ \  OR \ \ \overrightarrow{H_q{}_({}_2{}_)},\ \  ...,\ \  \overrightarrow{H_p{}_({}_i{}_)} \ \  OR \ \ \overrightarrow{H_q{}_({}_i{}_)}, \ \  ...,\ \  \overrightarrow{H_p{}_({}_d{}_)} \ \  OR \ \ \overrightarrow{H_q{}_({}_d{}_)}\} \label{equ:or}
\end{equation}

\end{figure*}

Figure \ref{search_overview} shows the details of the entire HDC executor process. After tokenization, we refer to the base hypervector generation process "Phase 1", while the process of encoding hypervectors and producing resultant hypervectors are referred to as "Phase 2" and "Phase 3", respectively. These 3 phases are the most important components to the process in the executor. Throughout the entire process, there are 10 kinds of hypervector properties or operations, which can produce different models. 

For "Phase 1", there are select properties for the base hypervector, including 2 global hypervector properties(\textcircled{1}\textcircled{2})(kept consistent during the whole process) and 1 local hypervector property (\textcircled{3}) (specific to the base hypervector). We can choose different hypervector dimensions for the base hypervector, such as 5000 and 10000. For traditional HDC, the sparsity (e.g., \# of 1) is fixed at 50\%. However, in the hyperdimensional computing architecture search, the sparsity is also a selectable parameter to produce different hypervectors to search suitable models for different scenarios. For the base hypervector, we may choose a binary or bipolar datatype, such as in traditional HDC.

For "Phase 1 to Phase 2", there are included 1 global property, 1 local property, and 3 local hypervector operations that can produce different encoded hypervectors. Each unit string will be broken down into substrings of length $N$, and this method is called $N-gram$. $Gram-size$(\textcircled{4}) is the number $N$ in the N-gram method, which treats $N$ continuous character(s) as a sub-unit that will be operated by element-wise operation(\textcircled{6}). Permutation (\textcircled{5}) is another optional choice in our method. The permutation is used to represent the appearance order of the characters. For example, in a string "that", the hypervectors which can represent the first "t" and the last "t" should not be the same but largely relative. So, rotating the base hypervector which corresponds to "t" for different elements is a good method. In traditional HDC, the rotation digits of elements are also relative to the number $N$ in the $N-gram$ method. The $i-th$ character in the string should rotate $N - i$ elements. For example, the first character in the string should rotate $N - 1$ digits. Since the base hypervectors are randomly generated, the rotation digits can vary. So, in our method, every hypervector can rotate $(N - i) * j$ elements, where $j$ is a non-negative integer. Specifically, if the $N$ in the N-gram method is larger than the length of the unit string $L$, the length of the unit string $L$ should be used as $N$ in the $N-gram$ method and it will be (just) applied for this unit string.
For hypervectors corresponding to $N$ continuous character(s), after rotation, an element-wise operation needs to be carried out. Since the element-wise operation is used to relate two hypervectors, we can have more choices. So, in our method, different from traditional HDC, 
there are 4 choices: 1) element-wise multiplication (donated as $*$), 2) element-wise XOR (donated as $XOR$), 3) element-wise AND (donated as $AND$), 4)element-wise OR (donated as $OR$), instead of only element-wise multiplication and XOR. Specifically, the 4 element-wise operations are described below. In these operations, we denote the first hypervector as $\overrightarrow{H_P}$ = \{ $\overrightarrow{H_p{}_({}_1{}_)}$, $\overrightarrow{H_p{}_({}_2{}_)}$, ..., $\overrightarrow{H_p{}_({}_i{}_)}$, ..., $\overrightarrow{H_p{}_({}_d{}_)}$\} and the second hypervector as  $\overrightarrow{H_Q}$ = \{ $\overrightarrow{H_q{}_({}_1{}_)}$, $\overrightarrow{H_q{}_({}_2{}_)}$, ..., $\overrightarrow{H_q{}_({}_i{}_)}$, ..., $\overrightarrow{H_q{}_({}_d{}_)}$\}. Equation \ref{equ:mul},\ref{equ:xor},\ref{equ:and},\ref{equ:or} show the 4 element-wise operation for 2 hypervectors.
Here, We re-define the element-wise operation results for bipolar data. Table \ref{tab:truth_table} shows the element-wise operation results for binary and bipolar data in our method.
\begin{table}[t]
  \centering
  \small
  \tabcolsep 4.8pt
  \caption{Truth table for binary and bipolar data when doing element-wise operation}
    \begin{tabular}{|c|c|c||c|c|c||c|c|c||c|c|c|}
    \hline
    \textbf{Mult.} & 1     & 0     & \textbf{XOR} & 1     & 0     & \textbf{AND} & 1     & 0     & \textbf{OR} & 1     & 0 \\
    \hline
    1     & 1     & 0     & 1     & 0     & 1     & 1     & 1     & 0     & 1     & 1     & 1 \\ \hline
    0     & 0     & 0     & 0     & 1     & 0     & 0     & 0     & 0     & 0     & 1     & 0 \\
    \hline
    \hline
    \textbf{Mult.} & 1     & -1    & \textbf{XOR} & 1     & -1    & \textbf{AND} & 1     & -1    & \textbf{OR} & 1     & -1 \\
    \hline
    1     & 1     & -1    & 1     & -1    & 1     & 1     & 1     & -1    & 1     & 1     & 1 \\ \hline
    -1    & -1    & 1     & -1    & 1     & -1    & -1    & -1    & -1    & -1    & 1     & -1 \\
    \hline
    \end{tabular}%
  \label{tab:truth_table}%
\end{table}%

After element operation, the mathematical operation (\textcircled{7}) is accumulating all the hypervectors resulting from the element-wise operation in one unit (e.g., "that"), the only choice is to add a threshold or not. The encoded hypervector also has a local property, encoded hypervector datatype. We can choose to represent the hypervector with a suitable datatype (\textcircled{8}). Here, different from the base hypervector, the encoded hypervector datatype has more choices than  binary and bipolar. After doing these, we can get encoded hypervectors.

For "Phase 2 to Phase 3", there are 1 local property and 1 local hypervector operation. The mathematical operation (\textcircled{9}) accumulates the hypervectors which corresponds to the same label, and the only choice is to add a threshold or not. Similar to the encoded hypervector, the resultant hypervector datatype(\textcircled{10}) has more choices. All the resultant hypervectors are stored in associative memory.

\subsection{Search Space}
\begin{table}[t]
  \centering
  \small
  \tabcolsep 3pt
  \caption{Proposed Search Space for Hyperdimensional Computing Architecture}
    \begin{tabular}{|p{40pt}|l|l|c|}
    \hline
    \multicolumn{2}{|c|}{\textbf{Computing Stage}} & \textbf{Properties or Operations} & \multicolumn{1}{c|}{\textbf{Flex.}} \\
    \hline
    \multirow{3}[0]{*}{Global HV} & All hypervectors & HV Dimension & \textbf{\checkmark} \\
          & All hypervectors & HV Sparsity  & \textbf{\checkmark} \\
          & Base \& Enc. HV & Gram-size & \textbf{\checkmark} \\
    \hline
    \multirow{2}[0]{*}{Local HV} & Encoding Type & Base HV Type & \textbf{\checkmark} \\
          \multirow{2}[0]{*}{(Datatype)}& Encoded Type & Encoded HV Type & \textbf{\checkmark} \\
           & Resultant Type & Resultant HV Type & \textbf{\checkmark} \\
    \hline
    \multirow{3}[0]{*}{Local HV} & \multirow{3}[0]{*}{Phase1 ->2 } & Permutaiton & \textbf{\checkmark} \\
          &       & Element-wise Operation & \textbf{\checkmark} \\
          (Operation) &       & Mathematical Operation & $\times$ \\
    \cline{2-4}
          & Phase2 ->3  & Mathematical Operation & $\times$ \\
    \hline
    \end{tabular}%
  \label{tab:selectable}%
\end{table}%

As described in Section 3.2, once different hypervector properties and operations are selected, the different HDC models can be generated with different datasets. Thus, these different hypervector properties and operations build the search space. Table \ref{tab:selectable} shows  whether the properties or the operations are selectable during the whole process. Since the only choice for both of the two mathematical operations is to add a threshold or not, and the datatypes of encoded hypervector and resultant hypervector are natural thresholds, we assume these are not selectable at this point. 

\begin{figure*}[t]
\begin{center}
\centerline{\includegraphics[width=7in]{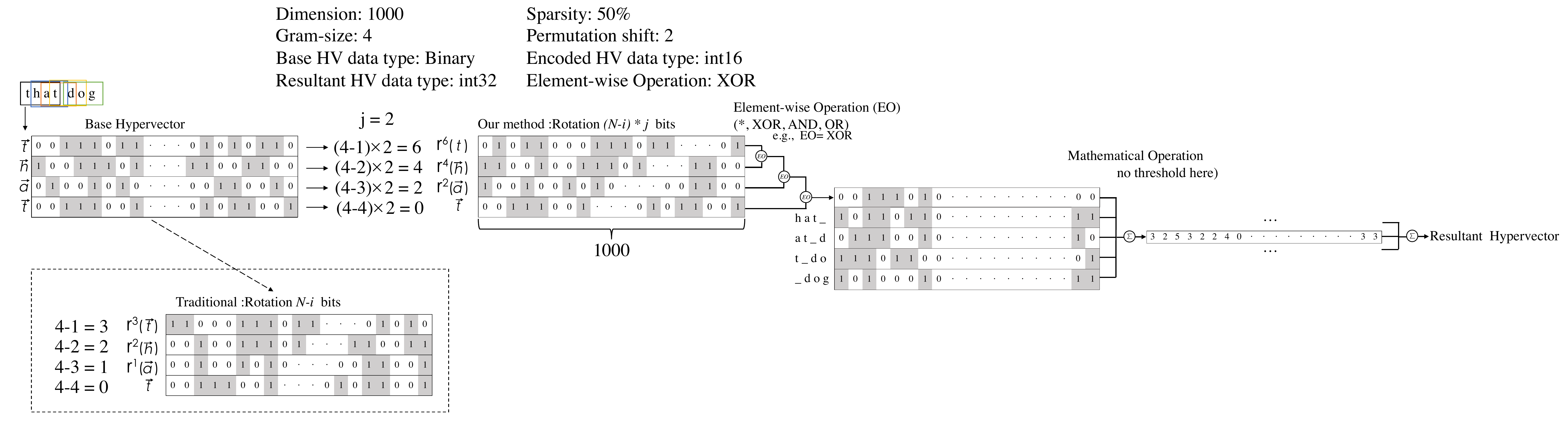}}
\vskip 0.1in
\caption{Example Process with Selected Parameters}
\label{example}
\end{center}
\vskip -0.2in
\end{figure*}

Figure \ref{example} better illustrates the process. After the RNN controller selected the parameters (properties and operations), the model is generated with the dataset. 
In this example, the dimension is 1000; the hypervector sparsity(\# of 1) is 50\%; the Gram-size is 4; the base hypervector datatype is binary; the encoded hypervector datatype is int16; the resultant hypervector datatype is int32; the shift of permutation is 2; the element-wise operation is XOR; no specific threshold in two mathematical operations. The example string is "that dog", and we use the first 4-gram substring to show the process. After tokenization, the string, whose length is 8, will be broken down into 5 substrings whose lengths are 4. These 5 substrings are "that", "hat\_", "at\_d", "t\_do", "\_dog" (here, "\_" is used to represent blank space).
The base hypervector corresponding to the character will rotate according to the character position in the sub-string. In order to show the permutation difference between our method and traditional HDC, the figure also shows the traditional permutation operation (shown in the dashed box). As introduced above, in our method, the hypervector needs to rotate $(N - i) * j$ elements, and in this example, the first $\overrightarrow{t}$ needs to rotate 6 elements, and the $\overrightarrow{h}$ and $\overrightarrow{a}$ needs to rotate 4 and 2 elements, respectively, while the last $\overrightarrow{t}$ does not need to rotate. 
After this, all of the 5 hypervector produced by 5 substrings need to be added together to generate an encoded hypervector. And all of the hypervectors with the same label will add together to generate the resultant hypervector.


\section{Experiments}\label{sec:exp}

\subsection{Experimental Setup}
\textbf{\textit{A. Dataset:}} We use a binary classification task from a drug discovery dataset and a 21-class classification task from 2 language datasets to evaluate the performance and compare it with other models. 
\begin{itemize}[noitemsep,topsep=0pt,parsep=0pt,partopsep=0pt]
    \item Clintox \cite{gayvert2016data}:The dataset includes two classification tasks for 1491 drug compounds with known binary chemical structures: (1) clinical trial toxicity and (2) FDA approval status. In our experiment, we concentrate on the task of clinical trial toxicity.
    \item Wortschatz Corpora \cite{quasthoff2006corpus}: The dataset contains 21 kinds of European languages with over 210,000 sample texts totally. 
    \item Europarl Parallel Corpus \cite{koehn2005europarl}: An independent text source contains the same 21 kinds of European languages with Wortschatz Corpora. This dataset provides 1,000 sample texts for each language, and each sample text is a single sentence. 
\end{itemize}
For the Clintox dataset, we used the scaffold split method to separate the datasets of 80\%, 10\%, and 10\% to build the training, validation, and testing sets, respectively. For the language recognition task, we used Wortschatz Corpora as the training set while using Europarl Parallel Corpus as the testing set. Since the  Wortschatz Corpora dataset is very large, we used approximately 21,000 sample texts, nearly 1/10 of the entire dataset, to speedup our AutoHDC  searching process. And in the searching process, we randomly split the dataset to 80\% and 20\% to build the training and validation set. After getting the models, the training process and retraining process used the whole dataset.

\textbf{\textit{B. AutoHDC settings:}}
In order to show the performance of hyperdimensional computing architecture search, we specified the search space as Table \ref{tab:Experimental_SS}. For HV Dimension, the search space increases from 1000 to 20000 with the step of 1000. For HV Sparsity, the percentage of the number of $1$ varies from 10\% to 90\% with the step of 10\%. For gram-size, $N$ of $N-gram$ varies from 1 to 6. For the base HV type, we can choose binary or bipolar. For encoded HV type and resultant HV type, we can choose binary, bipolar, int8, int16, int32, int64. For permutation, the shift can vary from 0 to 7 with a step of 1, and specifically, the HV will not rotate when the shift is 0. For element-wise operation, we can choose "*", "XOR", "AND" and "OR". For mathematical operation, we used the encoded datatype and the resultant datatype to be the thresholds instead of specifying other thresholds.

\begin{table*}[htbp]
  \centering
  \small
  \tabcolsep 8pt
  \caption{Search space utilized in the experimental evaluations}
    \begin{tabular}{|c|l|l|l|}
    \hline
          \multicolumn{2}{|c|}{\textbf{Computing Stage}}  & \textbf{Properties or Operations} & \multicolumn{1}{p{19em}|}{\textbf{Selectable Parameters}} \\
    \hline
    \multirow{3}{*}{Global HV Property} & All hypervectors & HV Dimension & (1k, 2k, …, 20k) \\
\cline{2-4}          & All hypervectors & HV Sparsity  & (10\%,20\%,…,90\%) \\
\cline{2-4}          & Base HV, Encoded HV & Gram-size & (1,2,3,4,5,6) \\
    \hline
    \multirow{3}{*}{Local HV Datatype} & Encoding Type & Base HV Type & (binary, bipolar) \\
\cline{2-4}          & Encoded Type & Encoded HV Type & (Binary, Bipolar, 8bit, 16bit, 32bit, 64bit) \\
\cline{2-4}          & Resultant Type & Resultant HV Type & (Binary, Bipolar, 8bit, 16bit, 32bit, 64bit) \\
    \hline
    \multirow{4}{*}{Local HV Operation} & \multirow{3}{*}{Phase1 -> 2 } & Permutaiton & (shift = 0,1,2,3,4,5,6,7) \\
\cline{3-4}          &       & Element-wise Operation & (*,XOR,AND,OR) \\
\cline{3-4}          &       & Mathematical Operation  & (+ (Datatype as Threshold)) \\
\cline{2-4}          & Phase2 -> 3  & Mathematical Operation & (+ (Datatype as Threshold)) \\
    \hline
    \end{tabular}%
  \label{tab:Experimental_SS}%
\end{table*}%

\textbf{\textit{C. Training settings:}} To evaluate the performance of  hyperdimensional computing architecture search, we select a set of state-of-the-art methods for comparison. For the Clintox dataset, it includes D-MPNN \cite{yang2019learned, swanson2019message}, MoleHD \cite{ma2021molehd} and 2 methods, Random Forest(RF) and Graph Convolutional Networks (GCN) from "MoleculeNet of deepchem" \cite{wu2018moleculenet,MoleculeNet}. For the language recognition task, it includes Random Indexing\cite{joshi2016language}, Robust and Energy-Efficient Classifier (REEC) \cite{rahimi2016robust}, HyperSeed \cite{osipov2021hyperseed} and SOMs \cite{kleyko2019distributed}.

For each model searched by our framework, it will run 5 times according to the randomly produced seed to random generate base HVs. For every model, we can get 5 scores (or accuracy for language recognition task), and the average score (accuracy) is used to be the reward of the RNN controller. For each dataset, the framework will run 500 episodes to search for the best models. In the searching process, models searched by our framework will be retrained only 10 epochs. The method in "MoleculeNet of deepchem" trained 1000 epochs. For a fair comparison, the models searched by our framework will be retrained 1000 epochs too. Since we found 1000 epochs may be too heavy for HDC models, so after every epoch, the model will do inference on the test set to get the performance to find the best model. Kindly note that the model would not change according to the test set.  


\textbf{\textit{D. Metrics:}} Drug discovery datasets are much more likely to be imbalanced, so accuracy is not a very commonly used metric to show the performance of models. Instead, receiver operating characteristics (ROC) curves and ROC Area-under-curve (AUC) scores are commonly used for binary and imbalanced datasets. Moreover, it is also suggested by benchmark datasets along with the majority of literature \cite{wu2018moleculenet, ramsundar2019deep}. However, it requires "probability" to compute the ROC-AUC score in neural network models. For HDC, the "similarity" computed through distance (e.g., \textit{cosine distance}) between resultant hypervector and query hypervector is used to predict. Similar to "probability", the largest "similarity" means the model predicts the query hypervector should belong to the class with the largest "similarity". So, in HDC, "similarity" is approximate "probability" in neural network models. Thus, we use the "similarity" to replace the "probability" in the ROC-AUC computation. Specifically, we use the \textit{softmax} function to let the summation of "similarity" be 1. The equation \ref{eq:sim} is shown below, where $\overrightarrow{Q}$ represents query hypervector,  $\overrightarrow{R}$ represents resultant hypervector, $P$ represents the probability. For the language recognition task, we still use the accuracy, which is the percentage of samples in the test (validation) set that are identified correctly.

\begin{equation}
    P = Softmax(S(\overrightarrow{Q}, \overrightarrow{R})) \label{eq:sim}
\end{equation}

\subsection{Experimental results}


\begin{table}[t]
  \centering
  \small
  \tabcolsep 12pt
  \caption{Exprimental results for drug discovery}
    \begin{tabular}{|l|c|c|}
    \hline
    \textbf{Models} & \textbf{Validation} & \textbf{Test} \\
    \hline
    \textbf{AutoHDC \_Clintox} & \textbf{0.9980 } & \textbf{0.9949 } \\
    \hline
    \textbf{MoleHD \cite{ma2021molehd}} & -     & 0.9870  \\
    \hline
    \textbf{MoleculeNet\_RF \cite{wu2018moleculenet, MoleculeNet}} & 0.8883  & 0.7829  \\
    \hline
    \textbf{MoleculeNet\_GCN \cite{wu2018moleculenet, MoleculeNet}} & 0.9880  & 0.9065  \\
    \hline
    \textbf{D-MPNN \cite{yang2019learned, swanson2019message}} & -     & 0.8740  \\
    \hline
    \end{tabular}%
  \label{tab:result_Clintox}%
\end{table}%


Table \ref{tab:result_Clintox} shows the ROC-AUC result of scaffold split Clintox test set processed by model searched by AutoHDC and competitors. 
Results show that our framework can find the best model for Clintox. The model searched by AutoHDC, called AutoHDC\_Clintox in Table \ref{tab:result_Clintox}, can achieve a ROC-AUC score of 0.9980 for the validation set and 0.9949 for the test set, which are the highest scores among those of competitors. For the validation set, compared with MoleculeNet\_RF and MoleculeNet\_GCN, AutoHDC\_Clintox improved the ROC-AUC score by 12.35\% at most. For the test set, compared with all competitors, AutoHDC\_Clintox can increase the ROC-AUC score up to 27.08\%.

\begin{table}[t]
  \centering
  \small
  \tabcolsep 12pt
  \caption{Exprimental results for language recognition}
    \begin{tabular}{|c|c|c|}
    \hline
    \textbf{Models} & \textbf{Validation} & \textbf{Test} \\
    \hline
    \textbf{AutoHDC\_Language} & 97.77\% & \textbf{99.07\%} \\
    \hline
    \textbf{Random Indexing \cite{joshi2016language}} & -     & 97.80\% \\
    \hline
    \textbf{REEC \cite{rahimi2016robust}} & -     & 97.10\% \\
    \hline
    \textbf{HyperSeed \cite{osipov2021hyperseed}} & -     & 91.00\% \\
    \hline
    \textbf{SOMs \cite{kleyko2019distributed}} & -     & 95.70\% \\
    \hline
    \end{tabular}%
  \label{tab:result_lang}%
\end{table}%

For the language recognition task, Table \ref{tab:result_lang} gives the results. AutoHDC\_Language (the model searched by AutoHDC for this task) achieves the 97.77\% of accuracy for the validation set. The accuracy is only 0.03\% lower than the test result of Random Indexing and higher than other test accuracy resulted from other competitors. But for the test set, AutoHDC\_Language achieved 99.07\% of accuracy, which is the highest among all competitors. In addition, AutoHDC\_Language can improve the test accuracy by at least 1.27\% and at most 8.07\%.

Combine all evaluation results, our proposed  AutoHDC has been verified to be effective to find the models with the best performance for Clintox and the language recognition task compared with competitors. 


\section{Conclusion}\label{sec:con}
We proposed an automated architecture search framework  to explore the architecture of brain-inspired hyperdimensional computing. To the best of our knowledge, this is the first paper exploring the HDC architecture and formulating the HDC architecture search space. Additionally, we apply reinforcement learning on top of our framework to make the search automatic.  Different from the current HDC, automated architecture search for HDC can search for suitable models for different datasets, which has shown competitive performance against deep neural networks in our test cases and may achieve comparable results more generally. The experimental results show that on a drug discovery dataset, Clintox, and a language recognition task (using two independent datasets), our framework can effectively search and find the good HDC models that achieve the best ROC-AUC scores and accuracy, with as much as 27.08\% improvement compared with alternate methods. These results suggest our approach to HDC architecture parallels that of the NAS, eliminating the need for ad-hoc approaches to adapt the methodology to specific scenarios.





\bibliography{CoExplore}

\end{document}